\documentclass{article}
\usepackage{algorithm,algpseudocode}
\usepackage{PRIMEarxiv}

\usepackage[utf8]{inputenc} % allow utf-8 input
\usepackage[T1]{fontenc}    % use 8-bit T1 fonts
\usepackage{hyperref}       % hyperlinks
\usepackage{url}            % simple URL typesetting
\usepackage{booktabs}       % professional-quality tables
\usepackage{amsfonts}       % blackboard math symbols
\usepackage{nicefrac}       % compact symbols for 1/2, etc.
\usepackage{microtype}      % microtypography
\usepackage{lipsum}
\usepackage{fancyhdr}       % header
\usepackage{graphicx}       % graphics
\graphicspath{{./}}     % organize your images and other figures under media/ folder

\usepackage{siunitx}
\usepackage{threeparttable}

\usepackage{sourcecodepro}
\usepackage{listings}
\usepackage{xcolor}
\usepackage{caption}

\title{AutoReason: Automatic Few-Shot Reasoning Decomposition
}

\author{
  Arda Sevinc, Abdurrahman Gumus \\
  Department of Electrical and Electronics Engineering \\
  Izmir Institute of Technology \\
  \texttt{arda@ardasevinc.com, abdurrahmangumus@iyte.edu.tr} \\
}

\begin{document}
\maketitle

\begin{abstract}
Chain of Thought (CoT) was introduced in recent research as a method
for improving step-by-step reasoning in Large Language Models.
However, CoT has limited applications such as its need
for hand-crafted few-shot exemplar prompts and no capability to adjust itself to different queries.

In this work, we propose a system to automatically generate rationales
using CoT. Our method improves multi-step implicit reasoning capabilities
by decomposing the implicit query into several explicit questions.
This provides interpretability for the model, improving reasoning in weaker LLMs.
We test our approach with two Q\&A datasets: StrategyQA and HotpotQA.
We show an increase in accuracy with both, especially on StrategyQA.

To facilitate further research in this field, the complete source code for this study has been made publicly available on GitHub: \url{https://github.com/miralab-ai/autoreason}.

\end{abstract}


\section{Introduction}
\label{sec:introduction}

The emergence of Large Language Models (LLMs) has marked a significant milestone
in the advancement of artificial intelligence and natural language processing \cite{radford_2019, raffel_2020, brown_2020jul}.
These powerful models, boasting billions of parameters and trained on massive amounts of text data, have demonstrated remarkable abilities in tasks such as language generation, question answering, and reasoning, surpassing human performance in some cases \cite{wang_2019, hendrycks_2021jan}.
The rapid progress in capabilities has sparked excitement and speculation
about their potential to enable more intelligent and human-like AI systems, with some researchers
even suggesting that they could be the key to achieving Artificial General Intelligence (AGI) \cite{bommasani_2022jul, wei_2022oct}.

Breakneck advancements in LLM capabilities have continued
with the introduction of GPT-4 level models, such as
Anthropic's Claude 3.5 Sonnet \cite{anthropic_2024a}, Claude 3 Opus \cite{anthropic_2024mar}, Google Deepmind's
Gemini 1.5 Pro \cite{GeminiUnlockingmultimodal_2024}, Llama 3 405b \cite{dubey_2024aug} and GPT4o \cite{gpt4o}.
These state-of-the-art models have demonstrated even more impressive
performance across a wide range of tasks, showcasing their potential
to revolutionize various industries and research domains.

Very recently ChatGPT o1-preview and o1-mini were released \cite{IntroducingOpenAIo1_2024sep},
emulating a system that looks like system II thinking.
Going from pattern recognition to analytical/critial thinking \cite{bonnefon_2018mar}.
Recent studies have further highlighted the remarkable abilities of these advanced models.
For instance, Bubeck et al. conducted a series of experiments on an early version of GPT-4, testing its
performance on a range of complex reasoning tasks. Their findings suggest
that GPT-4 exhibits "sparks of AGI," demonstrating the ability to solve problems that require abstract
reasoning, analogical thinking, and creative problem-solving.
These results underscore the potential of GPT-4 level models to push the boundaries of
what is possible with AI and to serve as powerful tools for advancing research
in areas such as natural language understanding, reasoning, and knowledge representation.

Reasoning is a crucial ability for language models, as it enables them
to draw inferences, make logical deductions, and solve complex problems \cite{bostrom_2018}.
However, despite their impressive performance on many natural
language tasks, LLMs still struggle with tasks that require multi-step reasoning
and the ability to combine multiple pieces of information \cite{rae_2022jan}.
This limitation hinders their potential to be used
in real-world applications that demand reliable and interpretable reasoning capabilities.

Chain of Thought (CoT) prompting \cite{wei_2023jan} has emerged as a promising
approach to address these limitations. By providing LLMs with examples
that include step-by-step reasoning traces, CoT prompting encourages
the models to generate similar traces for new problems, leading to improved
performance on reasoning tasks. However, the effectiveness of CoT prompting
heavily relies on the quality and relevance of the few-shot examples
used for prompting \cite{kojima_2023jan}. Crafting these examples manually
is time-consuming and requires significant expertise, limiting
the scalability and applicability of CoT prompting to new domains and tasks.

Despite the promising results of CoT prompting in enhancing
the reasoning capabilities of LLMs, current approaches suffer
from several limitations that hinder their scalability and applicability
to real-world scenarios. One major drawback is the reliance on manually
crafted few-shot examples, which require significant expertise and effort
to create \cite{kojima_2023jan}. This limitation makes
it challenging to apply CoT prompting to new domains
and tasks, as it demands the time-consuming process of designing high-quality, task-specific exemplars.

Moreover, existing CoT prompting methods typically use a fixed
set of exemplars for all queries, which may not always provide
the most relevant or informative reasoning traces for a given
problem \cite{wei_2023jan}. This lack of specificity can
lead to suboptimal performance and limit the ability of LLMs to adapt
their reasoning to the unique characteristics of each query.

In this work, we introduce AutoReason, a novel approach
that automatically generates rationales for each query
using CoT prompting. By generating these rationales, AutoReason
effectively transforms zero-shot queries into few-shot reasoning traces.
Which, in turn, is used by the system like CoT exemplars. Our main research questions are as follows:

\begin{enumerate}
    \item Can we increase the accuracy of zero-shot prompting by generating reasoning traces?
    \item Can we develop a method to automatically generate
    rationales using CoT and improve multi-step implicit reasoning in weaker LLMs?
    \item How can we generate unique rationales for each query, instead of
    relying on a fixed CoT prompt, to enhance the specificity and relevance of the reasoning traces?
    \item Can we demonstrate the effectiveness of automatically generated
    rationales in improving the reasoning performance of weaker LLMs on challenging multi-step reasoning tasks?
\end{enumerate}

By focusing on these questions, we seek to advance
the state of the art in LLM reasoning and make
CoT prompting more scalable and flexible. AutoReason distinguishes
itself from existing methods by generating rationales
from a zero-shot prompt automatically and tailoring
them to each specific query, thereby providing more relevant and informative reasoning traces.

The potential implications of our work are far-reaching.
By enabling the automatic generation of rationales
from a zero-shot prompt, AutoReason reduces barrier-to-entry of LLM prompting
and increases the surface area of chain-of-thought prompting to new domains
and tasks, making it more accessible and practical for real-world applications.
Moreover, by generating query-specific rationales, our approach has the potential to improve
the reasoning performance of LLMs on a wider range of problems, including those that require implicit, multi-step inference.

\subsection{Related Work}
\label{sec:related-work}

The development of increasingly powerful AI models
has highlighted the growing need for safe, interpretable, and reliable AI systems \cite{bostrom_2018}.
As these models become more capable, it is crucial to ensure that their
reasoning processes are transparent and understandable. Without interpretability, we risk having
intelligent systems that produce valid but uninterpretable answers, which can
limit their trustworthiness and hinder their application in real-world scenarios.

Interpretability is particularly important in the context of
language models, which have demonstrated remarkable performance
on a wide range of natural language tasks \cite{hendrycks_2021jan}.
However, as these models scale in size and complexity \cite{rae_2022jan}, their reasoning
processes become increasingly opaque, making it difficult to understand
how they arrive at their answers.
This lack of transparency can lead to unintended biases, errors, and potential misuse of these models.

To address these challenges, our work focuses on generating
intermediate reasoning steps that bridge the gap between
the input query and the final answer. By explicitly laying out these steps, we aim to make
the reasoning process of language models more interpretable and accessible to users.
This approach builds upon the foundation of Chain-of-Thought reasoning, which
has emerged as a promising technique for eliciting step-by-step explanations from language models.

\begin{figure}
  \centering
  \includegraphics[width=10cm]{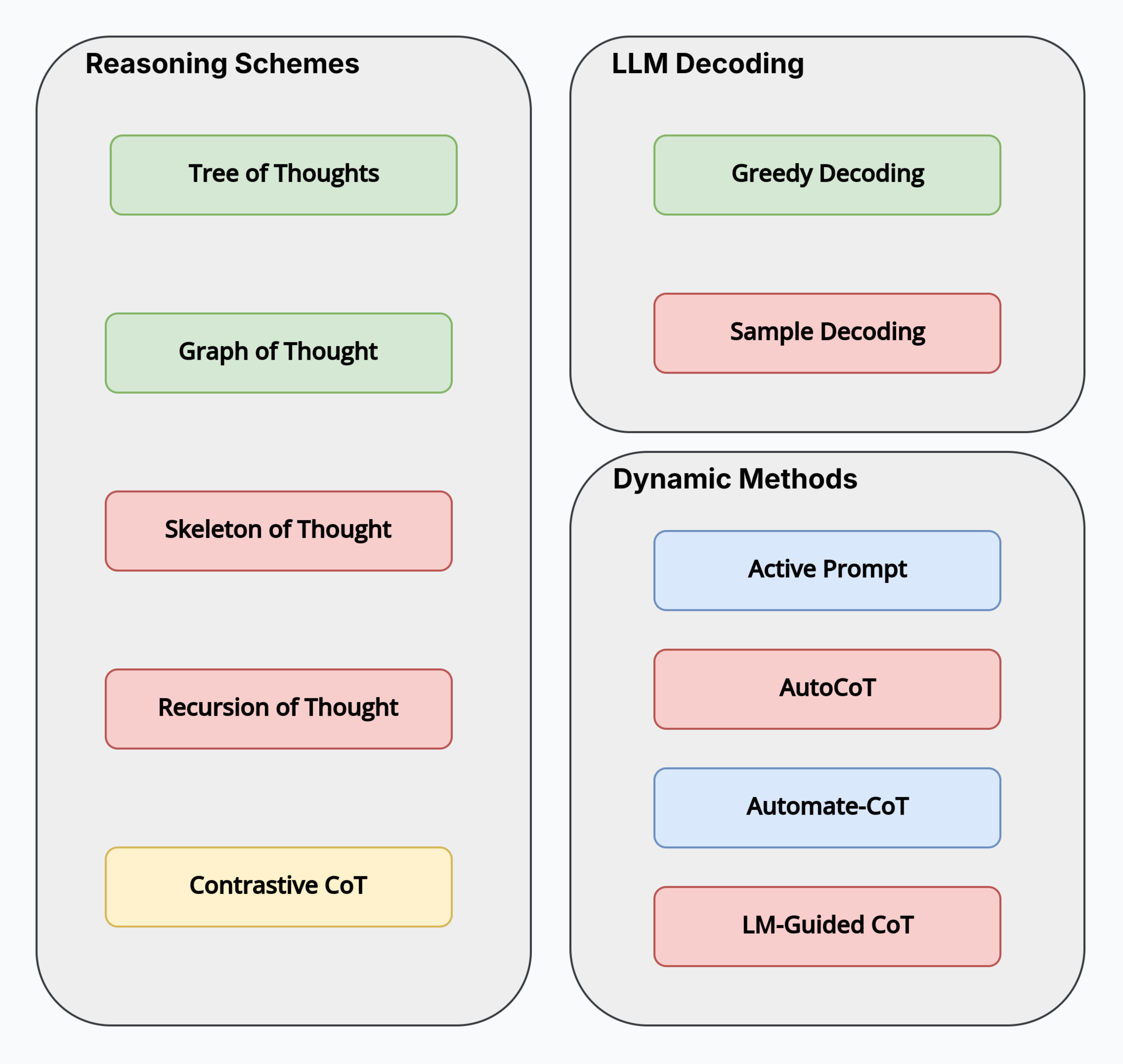}
  \caption{Reasoning methods based on Chain-of-Though (CoT).}
  \label{fig:cot-reasoning}
\end{figure}

CoT prompting \cite{wei_2023jan}, is a method for encouraging language
models to generate intermediate reasoning steps before providing a final answer.
By conditioning the model on a few examples that include step-by-step explanations, CoT prompting
has been shown to significantly improve the performance of language models
on a variety of reasoning tasks.
This approach has sparked a growing interest
in the research community, with several works exploring extensions and variations of the original CoT technique as can be seen in Figure \ref{fig:cot-reasoning}. 

Zero-Shot Chain-of-Thought \cite{kojima_2023jan} prompting is a method
designed to enhance the reasoning capabilities of large language models
by eliciting intermediate reasoning steps without requiring task-specific training examples.
As previously discussed, this is in contrast to normal chain-of-thought
where few-shot prompting is used.

\begin{figure}
  \centering
  \includegraphics[scale=0.343]{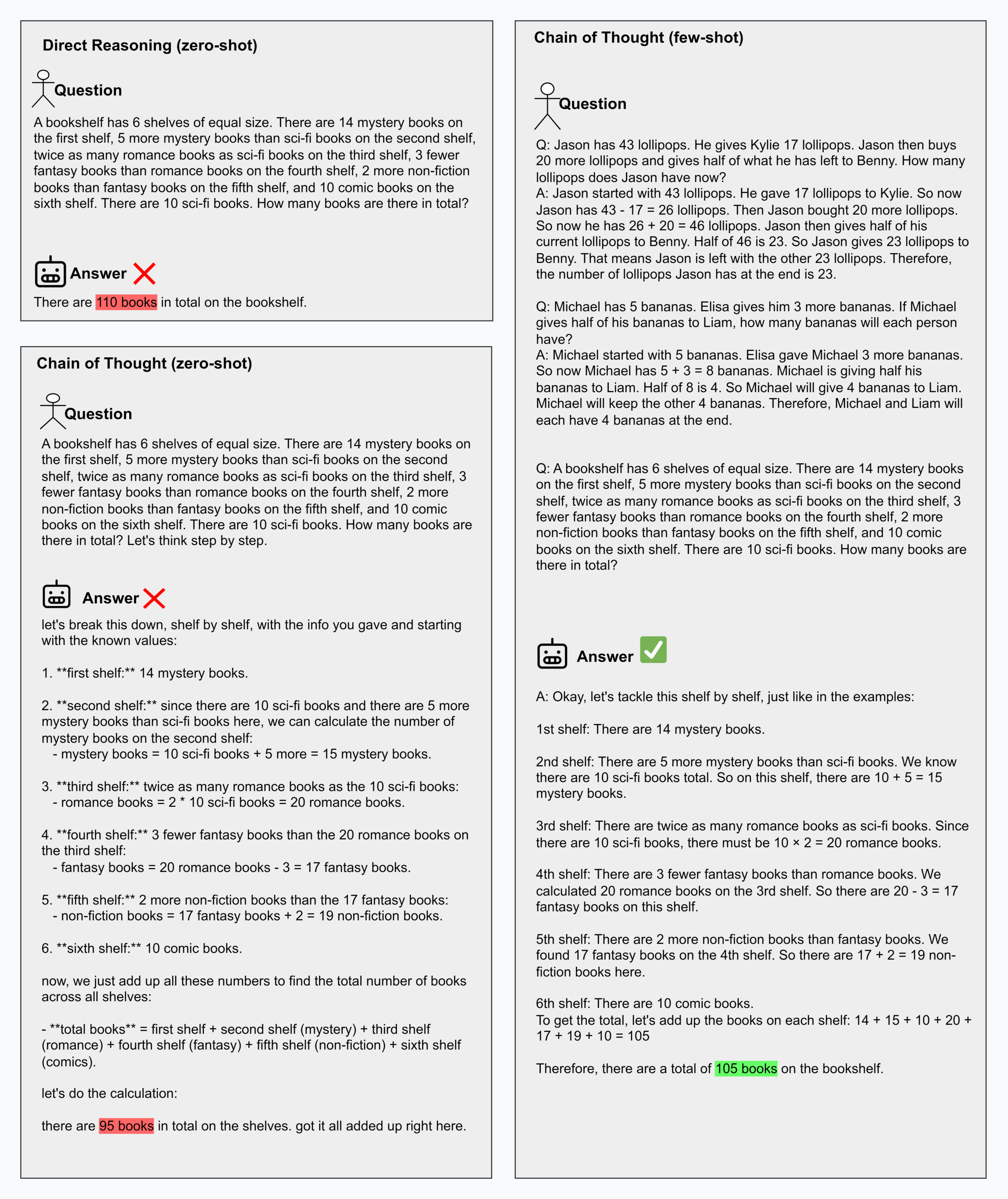}
  \caption{Few-Shot CoT and Zero-Shot CoT compared with direct Zero-Shot reasoning side by side.}
  \label{fig:cot-zero-few-shot}
\end{figure}

The key innovation of Zero-Shot CoT prompting is the use of a simple, task-agnostic
prompt such as "Let's think step by step" to guide the model to generate
a coherent series of reasoning steps leading to the final answer.
This method allows LLMs to tackle complex reasoning tasks by leveraging
their pre-existing knowledge and reasoning capabilities, providing a robust approach to zero-shot learning.
In Figure \ref{fig:cot-zero-few-shot} we see examples of direct reasoning, few-shot CoT and zero-shot CoT side by side.

Tree of Thoughts \cite{yao_2023may} is an advanced framework
developed to enhance the problem-solving capabilities of large language models
by structuring the reasoning process as a search over a tree of possible thought sequences.
Unlike traditional linear decision-making approaches, ToT allows LLMs to explore
multiple reasoning paths, evaluate potential outcomes, and iteratively choose the most promising path.
This method introduces two key strategies for
generating and evaluating thoughts: sampling diverse thoughts using
Chain-of-Thought prompts and proposing sequential thoughts tailored to the problem's constraints.

Graph of Thoughts \cite{besta_2023nov} is an advanced framework designed to enhance
the problem-solving capabilities of large language models by
representing the information they generate as a graph.
This approach allows for a more flexible and intricate form of reasoning
compared to linear or tree-based structures like Chain-of-Thought and Tree of Thoughts.
In GoT, each unit of information, or "thought," is a vertex, and the dependencies
between these thoughts are represented as edges.
This graph-based structure facilitates the combination, refinement, and generation of thoughts, enabling
the model to handle complex, multi-dimensional reasoning tasks more effectively.

Recursion of Thought \cite{lee_2023jun} is a novel framework designed to enhance the reasoning
capabilities of LLMs by leveraging a divide-and-conquer approach.
Inspired by human reasoning, RoT enables LLMs to recursively create
and utilize multiple contexts to solve complex problems that exceed
the model's maximum context size. This method introduces special tokens
that the models can output to trigger context-related operations, effectively
dividing the problem into smaller, manageable sub-problems and recursively solving them.

Skeleton of Thought \cite{ning_2023oct} is an innovative framework designed to reduce
generation latency in large language models by
implementing a parallel processing approach.
This technique involves generating an initial "skeleton" or outline
of the answer, which is then elaborated on in parallel, significantly
speeding up the inference process compared to traditional sequential decoding methods.
The process is divided into two main stages: the skeleton stage and the point-expanding stage.
During the skeleton stage, the model generates a concise outline of the answer.
In the point-expanding stage, the LLM expands on each point of the skeleton in parallel, which are then concatenated to form the final answer.

Program of Thoughts \cite{chen_2023octb} is an innovative prompting
framework designed to enhance the reasoning capabilities of large language models by
disentangling computation from reasoning, particularly for numerical and complex problem-solving tasks.
This method involves guiding the model through a structured sequence
of subtasks, where each subtask is addressed independently before combining the results to form the final answer.
This divide-and-conquer approach ensures that each part of the problem is solved efficiently and accurately,
reducing the cognitive load on the model and improving overall performance.

Self-consistency \cite{wang_2023mar} is a powerful framework designed to improve the robustness and accuracy of large language models by
generating multiple reasoning paths and consolidating them to form a final answer. Unlike traditional methods that rely on a single
chain of thought, self-consistency involves generating multiple
independent chains of thought for the same query, allowing
the model to explore various reasoning paths. These multiple outputs are then aggregated,
often through majority voting or other consensus mechanisms, to derive the most reliable final answer.
This approach leverages the diversity of the model's outputs to mitigate errors and enhance overall performance.

Contrastive Chain-of-Thought \cite{chia_2023nov} prompting is an advanced method
aimed at improving the reasoning capabilities of LLMs by providing both valid and invalid reasoning demonstrations.
This approach helps models learn more effectively by showing
them examples of both correct and incorrect reasoning, thus enabling them to understand
what mistakes to avoid. The contrastive CoT method leverages
this dual demonstration strategy to enhance the model's ability to generate accurate reasoning chains,
thereby improving performance on complex tasks such as arithmetic reasoning and factual question answering.

Active Prompt \cite{diao_2023may} is a novel method introduced to enhance the performance
of LLMs on complex reasoning tasks by leveraging task-specific example prompts annotated with CoT reasoning.
Unlike traditional CoT methods that rely on a fixed set of human-annotated exemplars, Active Prompt
dynamically selects the most uncertain questions for annotation using several
uncertainty metrics, such as disagreement, entropy, and variance.
This active selection process ensures that the annotated exemplars are the most
informative for the task at hand, significantly improving model performance.
In comparison, AutoReason focuses on decomposing zero-shot prompts into few-shot reasoning
traces, utilizing a stronger model (e.g., GPT-4) to generate detailed rationales
that a weaker model (e.g., GPT-3.5-turbo) can use to derive final answers.
While both methods aim to enhance reasoning capabilities through improved
prompt design, Active Prompt emphasizes the strategic selection and annotation
of uncertain questions, whereas AutoReason emphasizes the generation
of query-specific rationales to handle complex tasks.
Both approaches offer complementary insights into the optimization of CoT prompting for more intelligent and adaptable AI systems.

Another research exploring rationales is by \cite{wang_2022jul} Wang et al.'s paper
on rationale-augmented ensembles presents an innovative framework to enhance few-shot
learning in language models by leveraging rationale generation and ensemble techniques.
The key idea is to improve reasoning performance by generating
multiple rationales for each input and aggregating them to form a robust final output.
This approach includes self-consistency, prompt-order ensemble, and input-rationale
ensemble methods, which introduce diversity in generated rationales and enhance model
performance on complex reasoning tasks.
In comparison, AutoReason focuses on decomposing zero-shot prompts
into few-shot reasoning traces, using a stronger model (e.g., GPT-4) to generate
detailed rationales for a weaker model (e.g., GPT-3.5-turbo) to produce the final answers.
While both methods aim to improve reasoning capabilities, Wang et al.'s framework
emphasizes the importance of rationale diversity and aggregation through ensemble
techniques, whereas AutoReason prioritizes the generation of query-specific
rationales to enable weaker models to handle complex tasks.
Both methods provide valuable strategies for
optimizing reasoning in language models, highlighting the potential of rationale-based approaches in advancing AI capabilities.

Zhang et al. introduce Auto-CoT \cite{zhang_2022oct}, an automatic chain-of-thought
prompting method designed to eliminate the need for manual
demonstration construction in large language models (LLMs).
Auto-CoT leverages zero-shot CoT prompting with the "Let's think step by step" prompt
to generate reasoning chains for diverse, clustered questions, ensuring representative
and informative demonstrations.
This approach consistently matches or surpasses the performance
of manually designed CoT prompts across various reasoning tasks.
In comparison, AutoReason decomposes zero-shot prompts into few-shot
reasoning traces, using stronger models (e.g., GPT-4) to create detailed
rationales for weaker models (e.g., GPT-3.5-turbo) to generate final answers.
While both methods aim to enhance reasoning capabilities
through improved prompt design, Auto-CoT emphasizes automatic and diverse demonstration
generation, whereas AutoReason focuses on detailed rationale decomposition.
Both approaches contribute valuable strategies for advancing CoT prompting in LLMs.

In summary, AutoReason aims to address the key limitations
of current CoT prompting methods and unlock new possibilities
for scalable and flexible reasoning in language models.
Through our novel approach of automatic rationale generation, we strive to make a significant
contribution to the field of language model reasoning
and pave the way for more intelligent and adaptable AI systems.

\section{Methods}
\label{sec:methods}

\subsection{AutoReason}
\label{sec:autoreason}

AutoReason is a multi-step reasoning framework designed for Large Language Models (LLMs)
that effectively deconstructs zero-shot prompts
from users into few-shot reasoning traces, also known as rationales.
By utilizing these dynamically generated reasoning traces, AutoReason
significantly improves the accuracy of weaker
language models on questions that require complex reasoning.

\begin{figure}
  \centering
  \includegraphics[width=10cm]{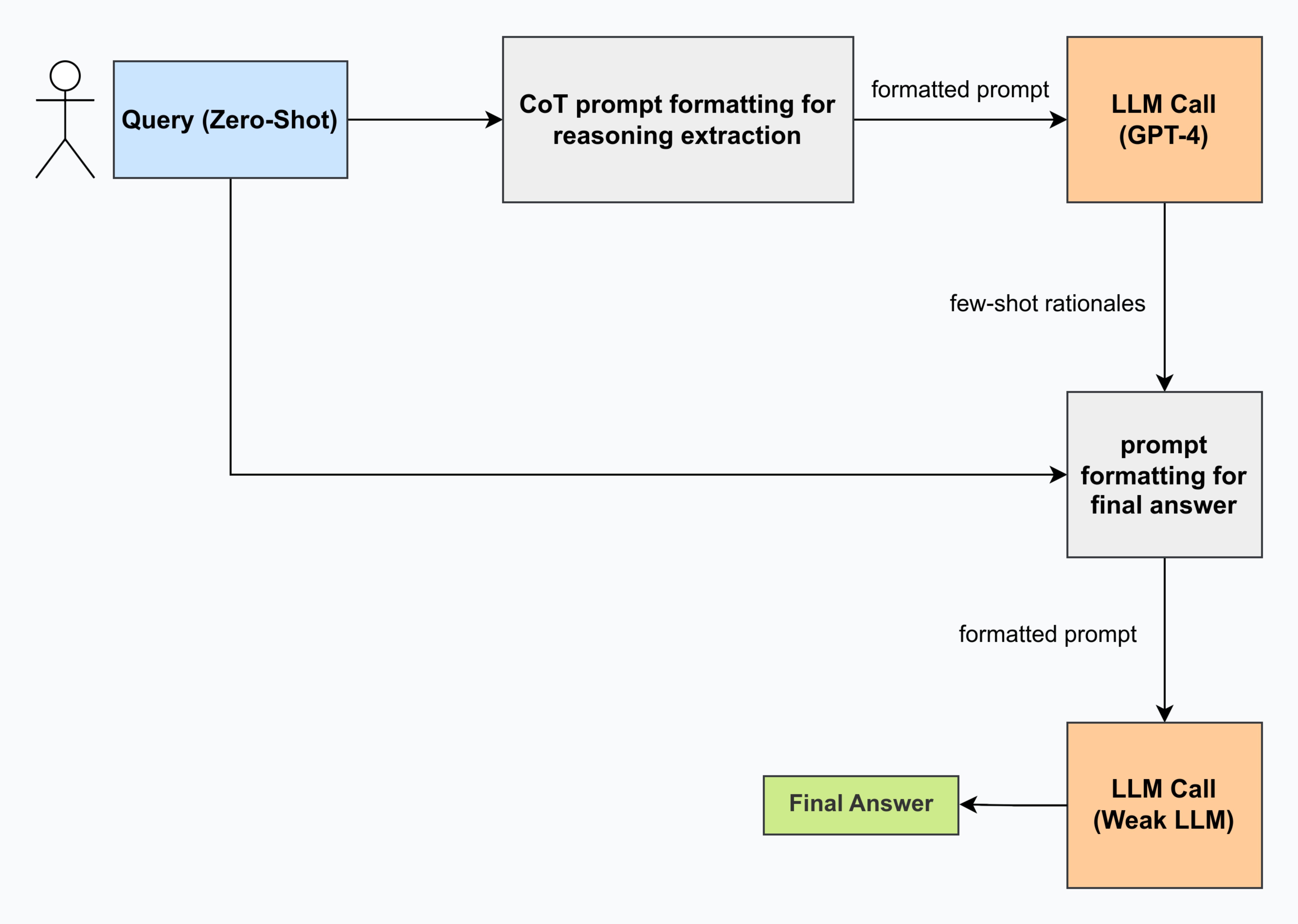}
  \caption{Block diagram illustrating the logical flow of prompt transformation of the AutoReason framework.}
  \label{fig:autoreason-framework}
\end{figure}

The AutoReason framework consists of several key components, as illustrated in Figure \ref{fig:autoreason-framework}
The initial query, which is assumed to be a zero-shot prompt, is first formatted
using a prompt template that includes several Chain-of-Thought (CoT) exemplars.
This carefully crafted prompt is designed to elicit rationales from the LLM by employing
CoT strategies, encouraging the model to break down
the problem into a series of explicit reasoning steps.
The complete prompt template can be found in the Appendix (Section A).

Once the generator prompt for reasoning extraction
is formatted, it is fed into GPT-4, a powerful LLM, through an API call to OpenAI.
GPT-4 then generates the rationales based on the provided prompt.
These rationales are subsequently formatted for obtaining the final answer
using another prompt template, which is also included in the Appendix.
By inserting both the initial query and the generated rationales
into this prompt, a weaker LLM, such as GPT-3.5-Turbo, is employed
to demonstrate the accuracy improvement achieved by AutoReason.

The modular and multi-stage approach of AutoReason ensures
interpretability and readability throughout the process
and evaluation steps, which will be discussed in detail
in the next section (2.2 Testing). After the final answer is obtained, it is scored
and classified according to the evaluation setup and testing methodology described in Section 2.2.

One of the key advantages of the AutoReason
framework is its adaptability to various LLMs by utilizing the provided
prompt templates. In our implementation, we chose GPT-4-1106-preview
for the rationale generator, leveraging its advanced capabilities
to decompose implicit reasoning into explicit rationales.
For demonstrating the effectiveness of the generated rationales
and obtaining the final answer, we employed GPT-3.5-Turbo-1106, a weaker LLM.

The pseudocode provided below shed light on how the devised algorithm works.

\begin{algorithm}[H]
\footnotesize
\caption{AutoReason Framework}
\begin{algorithmic}[1]
\Function{AutoReason}{query}
    \State formatted\_query $\gets$ \Call{FormatQueryWithCoTPrompt}{query}
    \State rationales $\gets$ \Call{GenerateRationalesWithGPT4}{formatted\_query}
    \State formatted\_prompt $\gets$ \Call{FormatPromptForFinalAnswer}{query, rationales}
    \State final\_answer $\gets$ \Call{GenerateFinalAnswerWithWeakerLLM}{formatted\_prompt}
    \State score $\gets$ \Call{ScoreAnswer}{query, final\_answer}
    \State \Return final\_answer, score
\EndFunction
\State
\Function{FormatQueryWithCoTPrompt}{query}
    \State cot\_prompt\_template $\gets$ \Call{LoadPromptTemplate}{"cot\_prompt.txt"}
    \State formatted\_query $\gets$ \Call{InsertQueryIntoTemplate}{query, cot\_prompt\_template}
    \State \Return formatted\_query
\EndFunction
\State
\Function{GenerateRationalesWithGPT4}{formatted\_query}
    \State response $\gets$ \Call{OpenAIApiCall}{"gpt-4-turbo-1106", formatted\_query}
    \State rationales $\gets$ \Call{ExtractRationales}{response}
    \State \Return rationales
\EndFunction
\State
\Function{FormatPromptForFinalAnswer}{query, rationales}
    \State final\_answer\_prompt\_template $\gets$ \Call{LoadPromptTemplate}{"final\_answer\_prompt.txt"}
    \State formatted\_prompt $\gets$ \Call{InsertQueryAndRationalesIntoTemplate}{query, rationales, final\_answer\_prompt\_template}
    \State \Return formatted\_prompt
\EndFunction
\State
\Function{GenerateFinalAnswerWithWeakerLLM}{formatted\_prompt}
    \State response $\gets$ \Call{OpenAIApiCall}{"gpt-3.5-turbo", formatted\_prompt}
    \State final\_answer $\gets$ \Call{ExtractFinalAnswer}{response}
    \State \Return final\_answer
\EndFunction
\State
\Function{ScoreAnswer}{query, final\_answer}
    \State correct\_answer $\gets$ \Call{LoadCorrectAnswer}{query}
    \State score $\gets$ \Call{CalculateScore}{final\_answer, correct\_answer}
    \State \Return score
\EndFunction
\end{algorithmic}
\end{algorithm}

The novelty of AutoReason lies in its two-step approach, which involves
rationale extraction followed by final answer generation. Although AutoReason
does not rely on dynamic CoT exemplars, the rationales
generated by GPT-4 are of high quality, enabling the framework
to effectively tackle implicit queries that require multi-step reasoning.
By decomposing complex implicit reasoning into a series of explicit reasoning
steps, AutoReason addresses the challenges faced
by language models when processing such queries, ultimately improving their accuracy and performance.

\subsection{Testing}
\label{sec:testing}

To evaluate the effectiveness of the AutoReason framework, we have developed
a comprehensive testing methodology that assesses
the accuracy of the generated answers. We focus on two datasets
specifically designed for multi-step reasoning tasks: HotpotQA and StrategyQA.

\begin{figure}
  \centering
  \includegraphics[width=15cm]{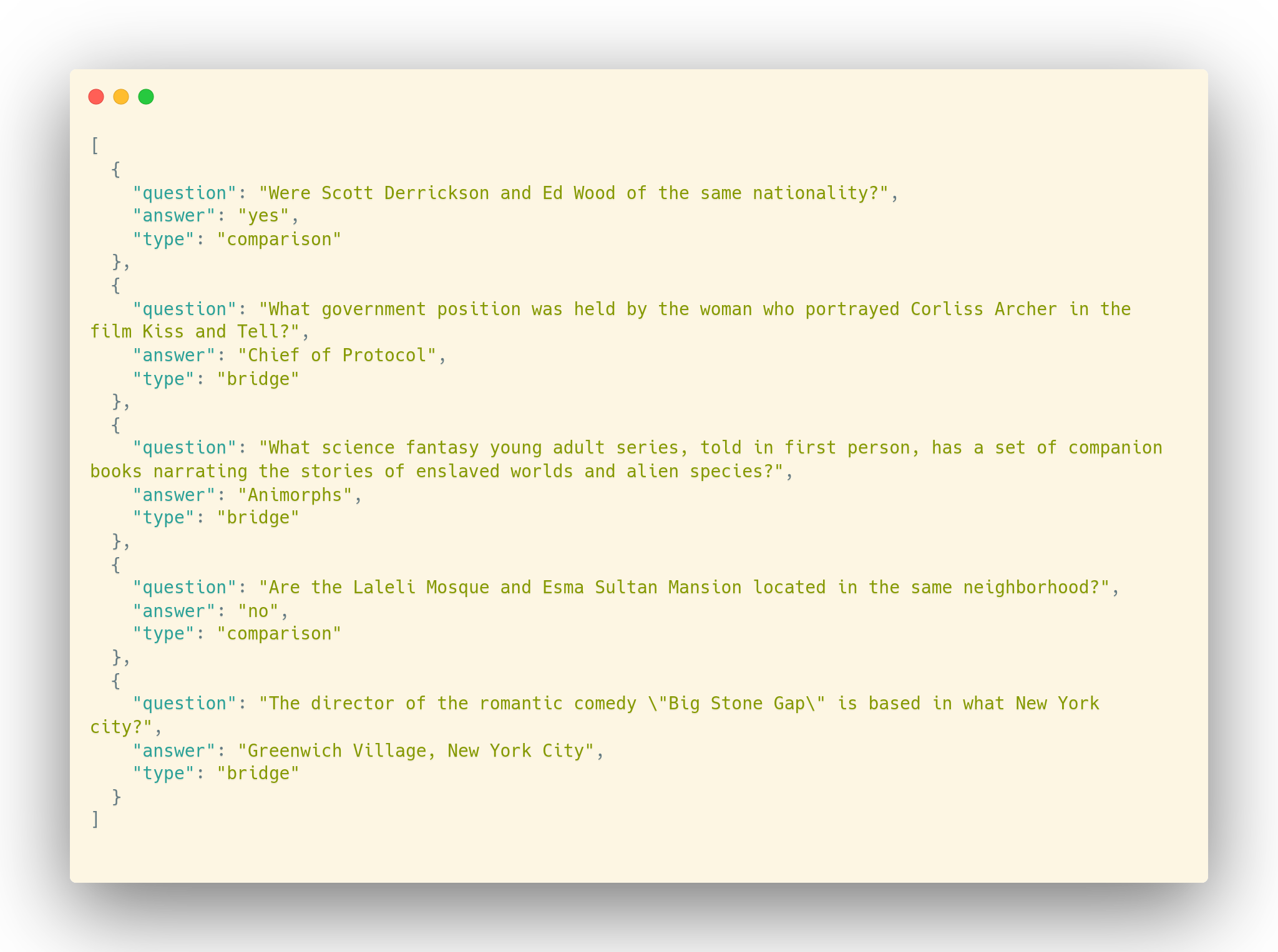}
  \caption{HotpotQA Example Data.}
  \label{fig:hotpotqa-example}
\end{figure}

HotpotQA \cite{yang_2018sep} is a dataset
containing over 7,000 question-answer pairs based on Wikipedia articles.
While HotpotQA aims to test multi-hop question answering, it is not particularly
well-suited for implicit reasoning tasks, which are the
primary focus of AutoReason.
As illustrated in Figure \ref{fig:hotpotqa-example}, HotpotQA
questions often require straightforward facts
to answer, rather than complex reasoning.
The impact of this characteristic on the results will be discussed in Section 3.

\begin{figure}
  \centering
  \includegraphics[width=10cm]{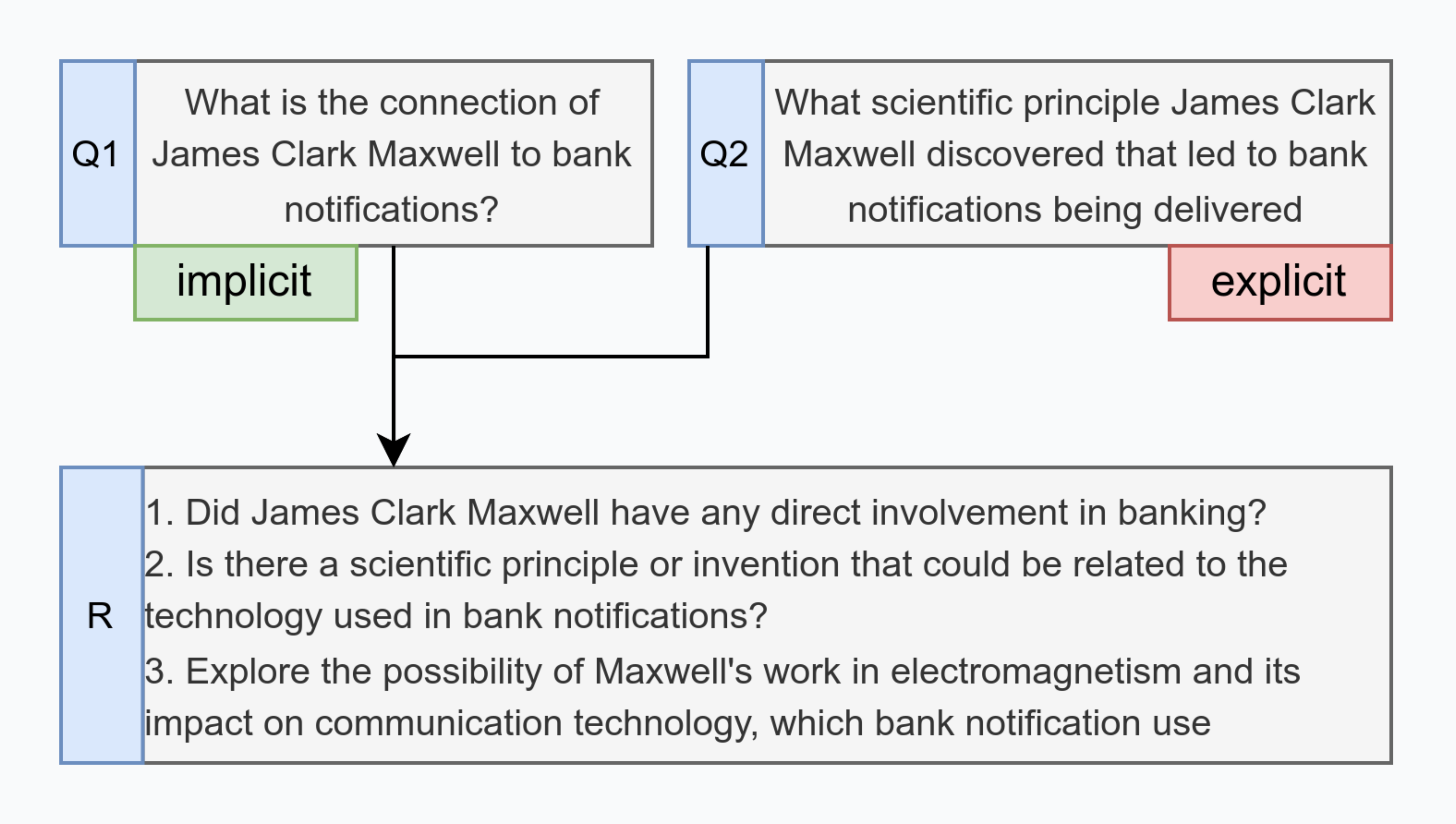}
  \caption{StrategyQA implicit reasoning.}
  \label{fig:strategyqa}
\end{figure}

In contrast, StrategyQA \cite{geva_2021} (Geva et al. 2021) is a human-curated
dataset with over 570 unique categories, specifically designed to test implicit
multi-step reasoning. The questions in StrategyQA
can only be answered by decomposing the problem
into a series of implicit reasoning steps, as exemplified
by the title question of the paper introducing the dataset: "Did Aristotle use a laptop?" (Geva et al. 2021).
Figure \ref{fig:strategyqa} demonstrates the process of decomposing
this question into a series of sub-questions that lead to the final answer.

To ensure the robustness and reliability of our evaluation, we employ the following testing setup:

\begin{enumerate}
    \item Shuffle the entire testing dataset using the Fisher-Yates algorithm.
    \item Sample a subset of the dataset with N=20 question-answer pairs.
    \item Test the sampled subset using the AutoReason framework.
    \item Score the generated answers using the methodology described in Section 2.1.
    \item Repeat steps 1-4 three times and calculate the average score across the three runs.
\end{enumerate}

\begin{figure}
  \centering
  \includegraphics[width=10cm]{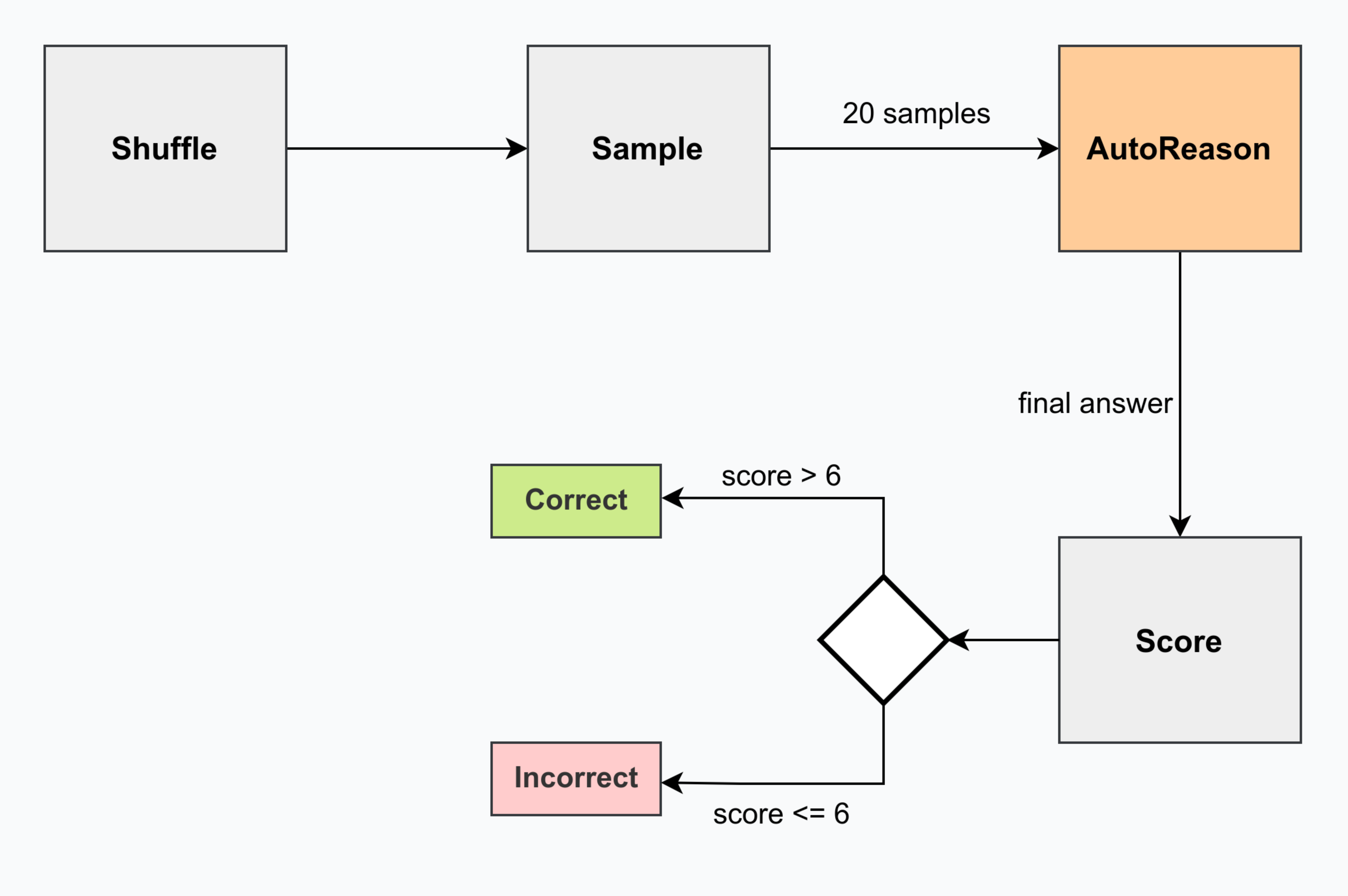}
  \caption{Block diagram exploring AutoReason's testing flow.}
  \label{fig:autoreason-testing}
\end{figure}

This testing flow is repeated three times, and the
final evaluation results are obtained
by averaging the scores across all three runs.
Scores are percentage of questions answered correctly according
to decision boundary.
Figure \ref{fig:autoreason-testing} provides a visual representation of this testing methodology.

Below is the the algorithm for the aforementioned testing setup.

\begin{algorithm}[H]
\footnotesize
\caption{Evaluation of the AutoReason Framework}
\begin{algorithmic}[1]
\Function{EvaluateAutoReason}{dataset, num\_samples, num\_runs, num\_iterations}
    \State scores $\gets$ []
    \For{$i \gets 1$ to num\_iterations}
        \State run\_scores $\gets$ []
        \For{$j \gets 1$ to num\_runs}
            \State shuffled\_dataset $\gets$ \Call{FisherYatesShuffle}{dataset}
            \State sampled\_dataset $\gets$ \Call{SampleDataset}{shuffled\_dataset, num\_samples}
            \State autoreason\_results $\gets$ \Call{TestAutoReason}{sampled\_dataset}
            \State run\_score $\gets$ \Call{ScoreAnswers}{autoreason\_results}
            \State run\_scores.\Call{Append}{run\_score}
        \EndFor
        \State iteration\_score $\gets$ \Call{CalculateAverage}{run\_scores}
        \State scores.\Call{Append}{iteration\_score}
    \EndFor
    \State final\_score $\gets$ \Call{CalculateAverage}{scores}
    \State \Return final\_score
\EndFunction
\end{algorithmic}
\end{algorithm}

By employing this rigorous testing setup, we aim to comprehensively
assess the performance of AutoReason on both HotpotQA and StrategyQA
datasets, providing insights into its effectiveness
in handling multi-step reasoning tasks and implicit reasoning challenges.

\section{Results and Discussion}
\label{sec:results-and-discussion}

\subsection{Results}
\label{sec:results}

The accuracy of the AutoReason framework was evaluated
on two datasets, HotpotQA and StrategyQA, using the testing
methodology described in section \ref{sec:testing}. The results, presented in Table \ref{tab:hotpotqa-results}
and Table \ref{tab:strategyqa-results}, demonstrate the effectiveness of our approach in
improving the reasoning capabilities of both weaker and stronger large language models.

\begin{table}[h!]
\centering
\begin{threeparttable}
    \caption{\raggedright \vspace{0.5em} AutoReason Testing Results on HotpotQA. All values are in percentages.}
    \begin{tabular*}{\textwidth}{@{\extracolsep{\fill}} l S[table-format=2.1] S[table-format=2.1] S[table-format=2.1]}
    \toprule
    \textbf{Model} & {\textbf{Base}} & {\textbf{CoT}} & {\textbf{AutoReason}} \\
    \midrule
    GPT-3.5-Turbo & 61.6 & 58.3 & 76.6 \\
    GPT-4-Turbo   & 73.3 & 63.3 & 71.6 \\
    \bottomrule
    \end{tabular*}
    \label{tab:hotpotqa-results}
\end{threeparttable}
\end{table}

On the StrategyQA dataset, which consists of questions
requiring implicit multi-step reasoning, AutoReason significantly
outperformed the baseline prompting models. GPT-3.5-Turbo, when used
with AutoReason achieved an accuracy of 76.6\%, surpassing
its base performance of 55\% and the CoT performance of 70.3\%. Similarly, GPT-4’s reasoning
accuracy increased form 71.6\% (base) to 76.6\% (CoT) to an impressive 91.6\% when using AutoReason.

However, on the HotpotQA dataset, which primarily
contains question answerable with straightforward facts, AutoReason’s performance
was mixed. GPT-3.5-Turbo’s accuracy increased on all
counts from 61.6\% (base) to 76.6\% on AutoReason, and surprisingly
decreased to 58.3\% on normal CoT. This highlights the superiority of our framework.

\begin{table}[h!]
\centering
\begin{threeparttable}
    \caption{\raggedright \vspace{0.5em} AutoReason Testing Results on StrategyQA. All values are in percentages.}
    \begin{tabular*}{\textwidth}{@{\extracolsep{\fill}} l S[table-format=2.1] S[table-format=2.1] S[table-format=2.1]}
    \toprule
    \textbf{Model} & {\textbf{Base}} & {\textbf{CoT}} & {\textbf{AutoReason}} \\
    \midrule
    GPT-3.5-Turbo & 55.0 & 70.3 & 76.6 \\
    GPT-4-Turbo   & 71.6 & 76.6 & 91.6 \\
    \bottomrule
    \end{tabular*}
    \label{tab:strategyqa-results}
\end{threeparttable}
\end{table}

Despite the expected result of increased accuracy, in GPT4,  we’ve noted a 1.7\% drop
from base prompting to AutoReason.
This regression is further noticed on
standard chain-of-thought, where the accuracy is observed to drop by 10\% to 63.4.
This observation or rather regression
highlights where in some cases, chain of thought
based prompting decreases accuracy as noted by Chen et. al. \cite{chen_2023octd}

In summary, AutoReason increased the accuracy
of both GPT-3.5-Turbo and GPT-4 compared to classic
chain-of-thought prompting and regular prompting except
in HotpotQA dataset, where a regression from base prompting
to AutoReason prompting was observed  on GPT4 - still with an increase compared to chain of thought.

\subsection{Discussion}
\label{sec:discussion}

The results of our study demonstrate the potential of AutoReason to enhance the reasoning capabilities of Large Language Models, particularly in tasks requiring complex, multi-step reasoning. However, these findings also reveal important nuances and limitations that warrant further discussion.

\subsubsection{Performance Across Datasets}

The divergent performance of AutoReason on StrategyQA and HotpotQA highlights the strengths and limitations of our approach. The significant improvement observed in StrategyQA tasks aligns with AutoReason's core design principle of decomposing implicit reasoning into explicit steps. StrategyQA questions, which often require intricate, multi-step reasoning that is not immediately apparent, benefit greatly from this decomposition process.

In contrast, the mixed results on HotpotQA suggest that AutoReason's benefits may be less pronounced for tasks that primarily rely on direct fact retrieval or simpler reasoning chains. This difference underscores the importance of matching reasoning enhancement techniques to the specific cognitive demands of different tasks.

\subsubsection{Model Behavior and Regression}

The observed regression in GPT-4's performance, particularly on HotpotQA, raises intriguing questions about the nature of LLM capabilities and their interaction with prompting techniques. While we lack definitive evidence, this regression may be indicative of increased model sophistication and sensitivity to prompts. As LLMs like GPT-4 evolve, they may develop a more nuanced understanding of query intent, sometimes leading to unexpected behaviors when presented with structured prompts designed for less advanced models.

This phenomenon highlights the dynamic nature of LLM development and the ongoing challenge of designing prompting strategies that remain effective as models become more advanced. It also underscores the need for continuous evaluation and adaptation of reasoning enhancement techniques like AutoReason.

\subsubsection{Implications for AGI and Complex Reasoning}

AutoReason's approach of using a stronger model to guide a weaker one in a two-step reasoning process bears similarities to recent developments in ``stage 2 thinking'' observed in models like OpenAI's o1 series. This parallel suggests that AutoReason may be tapping into fundamental principles of how advanced AI systems can approach complex reasoning tasks.

By demonstrating the potential for LLMs to engage in more deliberate, step-by-step reasoning processes, AutoReason contributes to the broader goal of developing Artificial General Intelligence (AGI). The ability to break down complex problems into manageable steps and reason through them systematically is a key aspect of general intelligence. However, AutoReason also highlights current limitations in LLM reasoning, particularly in maintaining consistency across long chains of thought and in handling tasks that require genuine causal understanding or abstract reasoning.

\subsubsection{Ethical Considerations and Societal Impact}

The development of systems like AutoReason, which aim to enhance the reasoning capabilities of AI, raises important ethical considerations. As these systems become more sophisticated, there is a risk of over-reliance on machine-generated rationales, potentially leading to the automation of decision-making processes in sensitive domains without adequate human oversight.

Moreover, as reasoning chains become more complex, there is a growing challenge of interpretability. If AI systems develop ways of communicating or reasoning that are not easily understood by humans, it could lead to a ``black box'' problem in critical reasoning tasks. This lack of transparency could have significant implications in fields such as healthcare, law, and finance, where the ability to explain and justify decisions is crucial.

\subsubsection{Limitations and Future Work}

While AutoReason shows promise, it is important to acknowledge its limitations. The quality of the generated rationales is crucial to the success of the method, and poor-quality rationales can lead to incorrect answers or hallucinations. This dependency on rationale quality highlights the need for robust evaluation metrics and quality control mechanisms in future iterations of the system.

The computational cost of using two LLMs in sequence, while not prohibitive with current API services, may become a consideration in large-scale applications. Future work should explore optimizations to improve efficiency without sacrificing reasoning quality.

Additionally, the current study's limited sample size and number of runs point to the need for more extensive testing across a broader range of tasks and domains. Expanding the evaluation to include diverse reasoning tasks beyond question-answering could provide valuable insights into the generalizability of AutoReason.

Future research directions could include:
\begin{enumerate}
    \item Investigating the integration of AutoReason with other AI techniques such as reinforcement learning or neuro-symbolic approaches.
    \item Exploring ways to make the reasoning process more transparent and interpretable.
    \item Developing methods to dynamically adjust the level of reasoning decomposition based on task complexity.
    \item Conducting user studies to assess the practical impact of AutoReason in real-world applications.
\end{enumerate}

In conclusion, AutoReason represents a step forward in enhancing the reasoning capabilities of LLMs, but it also illuminates the complexities and challenges inherent in developing AI systems capable of human-like reasoning. As we continue to refine and expand this approach, careful consideration of its implications and limitations will be crucial in realizing its full potential while mitigating potential risks.

\section{Conclusion}

This paper introduces AutoReason, a novel framework designed to enhance the reasoning capabilities of Large Language Models (LLMs) through automatic generation of reasoning traces. By leveraging a two-step process that combines the strengths of different LLMs, AutoReason demonstrates significant potential in improving performance on complex reasoning tasks, particularly those requiring implicit multi-step reasoning.

Our experimental results on the StrategyQA and HotpotQA datasets highlight both the strengths and limitations of AutoReason. The framework showed marked improvement in tasks requiring intricate, multi-step reasoning, as evidenced by the performance boost on StrategyQA. However, the mixed results on HotpotQA underscore the importance of aligning reasoning enhancement techniques with the specific cognitive demands of different tasks.

Key contributions of this work include:

\begin{enumerate}
    \item The development of a two-tier model approach that uses a stronger LLM to generate reasoning traces for a weaker LLM, effectively guiding its decision-making process.
    \item Demonstration of improved performance on complex reasoning tasks, particularly those involving implicit reasoning steps.
    \item Insights into the interaction between advanced LLMs and structured prompting techniques, including observations on model behavior and potential regressions.
    \item A framework that contributes to the broader goal of developing more robust and interpretable AI reasoning systems.
\end{enumerate}

Despite these advancements, AutoReason also reveals important challenges in the field of AI reasoning. The quality dependency of generated rationales, computational costs of using multiple LLMs, and the need for more extensive testing across diverse tasks are areas that require further investigation.

Looking forward, AutoReason opens up several promising avenues for future research:

\begin{enumerate}
    \item Integration with other AI techniques such as reinforcement learning or neuro-symbolic approaches to further enhance reasoning capabilities.
    \item Development of methods to improve the transparency and interpretability of the reasoning process.
    \item Exploration of dynamic reasoning decomposition techniques that adapt to varying task complexities.
    \item Investigation of AutoReason's potential in real-world applications through comprehensive user studies.
\end{enumerate}

In conclusion, while AutoReason represents a important step towards enhancing the reasoning capabilities of LLMs, it also illuminates the complexities involved in developing AI systems capable of human-like reasoning. As we continue to refine and expand this approach, careful consideration of its implications, limitations, and ethical considerations will be crucial in realizing its full potential while mitigating potential risks. The journey towards more advanced AI reasoning systems is ongoing, and AutoReason contributes an important piece to this evolving puzzle.

\bibliographystyle{unsrt}  
\bibliography{autoreason}  

\begin{thebibliography}{10}

\bibitem{radford_2019}
Alec Radford, Jeffrey Wu, Rewon Child, David Luan, Dario Amodei, and Ilya Sutskever.
\newblock Language {{Models}} are {{Unsupervised Multitask Learners}}.
\newblock 1(8):9, 2019.

\bibitem{raffel_2020}
Colin Raffel, Noam Shazeer, Adam Roberts, Katherine Lee, Sharan Narang, Michael Matena, Yanqi Zhou, Wei Li, and Peter~J. Liu.
\newblock Exploring the {{Limits}} of {{Transfer Learning}} with a {{Unified Text-to-Text Transformer}}.
\newblock {\em Journal of Machine Learning Research}, 21(140):1--67, 2020.

\bibitem{brown_2020jul}
Tom~B. Brown, Benjamin Mann, Nick Ryder, Melanie Subbiah, Jared Kaplan, Prafulla Dhariwal, Arvind Neelakantan, Pranav Shyam, Girish Sastry, Amanda Askell, Sandhini Agarwal, Ariel {Herbert-Voss}, Gretchen Krueger, Tom Henighan, Rewon Child, Aditya Ramesh, Daniel~M. Ziegler, Jeffrey Wu, Clemens Winter, Christopher Hesse, Mark Chen, Eric Sigler, Mateusz Litwin, Scott Gray, Benjamin Chess, Jack Clark, Christopher Berner, Sam McCandlish, Alec Radford, Ilya Sutskever, and Dario Amodei.
\newblock Language {{Models}} are {{Few-Shot Learners}}, July 2020.

\bibitem{wang_2019}
Alex Wang, Yada Pruksachatkun, Nikita Nangia, Amanpreet Singh, Julian Michael, Felix Hill, Omer Levy, and Samuel Bowman.
\newblock {{SuperGLUE}}: {{A Stickier Benchmark}} for {{General-Purpose Language Understanding Systems}}.
\newblock In {\em Advances in {{Neural Information Processing Systems}}}, volume~32. Curran Associates, Inc., 2019.

\bibitem{hendrycks_2021jan}
Dan Hendrycks, Collin Burns, Steven Basart, Andy Zou, Mantas Mazeika, Dawn Song, and Jacob Steinhardt.
\newblock Measuring {{Massive Multitask Language Understanding}}, January 2021.

\bibitem{bommasani_2022jul}
Rishi Bommasani, Drew~A. Hudson, Ehsan Adeli, Russ Altman, Simran Arora, Sydney {von Arx}, Michael~S. Bernstein, Jeannette Bohg, Antoine Bosselut, Emma Brunskill, Erik Brynjolfsson, Shyamal Buch, Dallas Card, Rodrigo Castellon, Niladri Chatterji, Annie Chen, Kathleen Creel, Jared~Quincy Davis, Dora Demszky, Chris Donahue, Moussa Doumbouya, Esin Durmus, Stefano Ermon, John Etchemendy, Kawin Ethayarajh, Li~{Fei-Fei}, Chelsea Finn, Trevor Gale, Lauren Gillespie, Karan Goel, Noah Goodman, Shelby Grossman, Neel Guha, Tatsunori Hashimoto, Peter Henderson, John Hewitt, Daniel~E. Ho, Jenny Hong, Kyle Hsu, Jing Huang, Thomas Icard, Saahil Jain, Dan Jurafsky, Pratyusha Kalluri, Siddharth Karamcheti, Geoff Keeling, Fereshte Khani, Omar Khattab, Pang~Wei Koh, Mark Krass, Ranjay Krishna, Rohith Kuditipudi, Ananya Kumar, Faisal Ladhak, Mina Lee, Tony Lee, Jure Leskovec, Isabelle Levent, Xiang~Lisa Li, Xuechen Li, Tengyu Ma, Ali Malik, Christopher~D. Manning, Suvir Mirchandani, Eric Mitchell, Zanele Munyikwa, Suraj Nair,
  Avanika Narayan, Deepak Narayanan, Ben Newman, Allen Nie, Juan~Carlos Niebles, Hamed Nilforoshan, Julian Nyarko, Giray Ogut, Laurel Orr, Isabel Papadimitriou, Joon~Sung Park, Chris Piech, Eva Portelance, Christopher Potts, Aditi Raghunathan, Rob Reich, Hongyu Ren, Frieda Rong, Yusuf Roohani, Camilo Ruiz, Jack Ryan, Christopher R{\'e}, Dorsa Sadigh, Shiori Sagawa, Keshav Santhanam, Andy Shih, Krishnan Srinivasan, Alex Tamkin, Rohan Taori, Armin~W. Thomas, Florian Tram{\`e}r, Rose~E. Wang, William Wang, Bohan Wu, Jiajun Wu, Yuhuai Wu, Sang~Michael Xie, Michihiro Yasunaga, Jiaxuan You, Matei Zaharia, Michael Zhang, Tianyi Zhang, Xikun Zhang, Yuhui Zhang, Lucia Zheng, Kaitlyn Zhou, and Percy Liang.
\newblock On the {{Opportunities}} and {{Risks}} of {{Foundation Models}}, July 2022.

\bibitem{wei_2022oct}
Jason Wei, Yi~Tay, Rishi Bommasani, Colin Raffel, Barret Zoph, Sebastian Borgeaud, Dani Yogatama, Maarten Bosma, Denny Zhou, Donald Metzler, Ed~H. Chi, Tatsunori Hashimoto, Oriol Vinyals, Percy Liang, Jeff Dean, and William Fedus.
\newblock Emergent {{Abilities}} of {{Large Language Models}}, October 2022.

\bibitem{anthropic_2024a}
Anthropic.
\newblock Introducing {{Claude}} 3.5 {{Sonnet}}.
\newblock https://www.anthropic.com/news/claude-3-5-sonnet.

\bibitem{anthropic_2024mar}
Anthropic.
\newblock The {{Claude}} 3 {{Model Family}}: {{Opus}}, {{Sonnet}}, {{Haiku}}, March 2024.

\bibitem{GeminiUnlockingmultimodal_2024}
{Gemini Team, Google}.
\newblock Gemini 1.5: {{Unlocking}} multimodal understanding across millions of tokens of context.
\newblock Technical report, Google DeepMind, 2024.

\bibitem{dubey_2024aug}
Abhimanyu Dubey, Abhinav Jauhri, Abhinav Pandey, Abhishek Kadian, Ahmad {Al-Dahle}, Aiesha Letman, Akhil Mathur, Alan Schelten, Amy Yang, Angela Fan, Anirudh Goyal, Anthony Hartshorn, Aobo Yang, Archi Mitra, Archie Sravankumar, Artem Korenev, Arthur Hinsvark, Arun Rao, Aston Zhang, Aurelien Rodriguez, Austen Gregerson, Ava Spataru, Baptiste Roziere, Bethany Biron, Binh Tang, Bobbie Chern, Charlotte Caucheteux, Chaya Nayak, Chloe Bi, Chris Marra, Chris McConnell, Christian Keller, Christophe Touret, Chunyang Wu, Corinne Wong, Cristian~Canton Ferrer, Cyrus Nikolaidis, Damien Allonsius, Daniel Song, Danielle Pintz, Danny Livshits, David Esiobu, Dhruv Choudhary, Dhruv Mahajan, Diego {Garcia-Olano}, Diego Perino, Dieuwke Hupkes, Egor Lakomkin, Ehab AlBadawy, Elina Lobanova, Emily Dinan, Eric~Michael Smith, Filip Radenovic, Frank Zhang, Gabriel Synnaeve, Gabrielle Lee, Georgia~Lewis Anderson, Graeme Nail, Gregoire Mialon, Guan Pang, Guillem Cucurell, Hailey Nguyen, Hannah Korevaar, Hu~Xu, Hugo Touvron, Iliyan
  Zarov, Imanol~Arrieta Ibarra, Isabel Kloumann, Ishan Misra, Ivan Evtimov, Jade Copet, Jaewon Lee, Jan Geffert, Jana Vranes, Jason Park, Jay Mahadeokar, Jeet Shah, Jelmer {van der Linde}, Jennifer Billock, Jenny Hong, Jenya Lee, Jeremy Fu, Jianfeng Chi, Jianyu Huang, Jiawen Liu, Jie Wang, Jiecao Yu, Joanna Bitton, Joe Spisak, Jongsoo Park, Joseph Rocca, Joshua Johnstun, Joshua Saxe, Junteng Jia, Kalyan~Vasuden Alwala, Kartikeya Upasani, Kate Plawiak, Ke~Li, Kenneth Heafield, Kevin Stone, Khalid {El-Arini}, Krithika Iyer, Kshitiz Malik, Kuenley Chiu, Kunal Bhalla, Lauren {Rantala-Yeary}, Laurens {van der Maaten}, Lawrence Chen, Liang Tan, Liz Jenkins, Louis Martin, Lovish Madaan, Lubo Malo, Lukas Blecher, Lukas Landzaat, Luke {de Oliveira}, Madeline Muzzi, Mahesh Pasupuleti, Mannat Singh, Manohar Paluri, Marcin Kardas, Mathew Oldham, Mathieu Rita, Maya Pavlova, Melanie Kambadur, Mike Lewis, Min Si, Mitesh~Kumar Singh, Mona Hassan, Naman Goyal, Narjes Torabi, Nikolay Bashlykov, Nikolay Bogoychev, Niladri
  Chatterji, Olivier Duchenne, Onur {\c C}elebi, Patrick Alrassy, Pengchuan Zhang, Pengwei Li, Petar Vasic, Peter Weng, Prajjwal Bhargava, Pratik Dubal, Praveen Krishnan, Punit~Singh Koura, Puxin Xu, Qing He, Qingxiao Dong, Ragavan Srinivasan, Raj Ganapathy, Ramon Calderer, Ricardo~Silveira Cabral, Robert Stojnic, Roberta Raileanu, Rohit Girdhar, Rohit Patel, Romain Sauvestre, Ronnie Polidoro, Roshan Sumbaly, Ross Taylor, Ruan Silva, Rui Hou, Rui Wang, Saghar Hosseini, Sahana Chennabasappa, Sanjay Singh, Sean Bell, Seohyun~Sonia Kim, Sergey Edunov, Shaoliang Nie, Sharan Narang, Sharath Raparthy, Sheng Shen, Shengye Wan, Shruti Bhosale, Shun Zhang, Simon Vandenhende, Soumya Batra, Spencer Whitman, Sten Sootla, Stephane Collot, Suchin Gururangan, Sydney Borodinsky, Tamar Herman, Tara Fowler, Tarek Sheasha, Thomas Georgiou, Thomas Scialom, Tobias Speckbacher, Todor Mihaylov, Tong Xiao, Ujjwal Karn, Vedanuj Goswami, Vibhor Gupta, Vignesh Ramanathan, Viktor Kerkez, Vincent Gonguet, Virginie Do, Vish Vogeti, Vladan
  Petrovic, Weiwei Chu, Wenhan Xiong, Wenyin Fu, Whitney Meers, Xavier Martinet, Xiaodong Wang, Xiaoqing~Ellen Tan, Xinfeng Xie, Xuchao Jia, Xuewei Wang, Yaelle Goldschlag, Yashesh Gaur, Yasmine Babaei, Yi~Wen, Yiwen Song, Yuchen Zhang, Yue Li, Yuning Mao, Zacharie~Delpierre Coudert, Zheng Yan, Zhengxing Chen, Zoe Papakipos, Aaditya Singh, Aaron Grattafiori, Abha Jain, Adam Kelsey, Adam Shajnfeld, Adithya Gangidi, Adolfo Victoria, Ahuva Goldstand, Ajay Menon, Ajay Sharma, Alex Boesenberg, Alex Vaughan, Alexei Baevski, Allie Feinstein, Amanda Kallet, Amit Sangani, Anam Yunus, Andrei Lupu, Andres Alvarado, Andrew Caples, Andrew Gu, Andrew Ho, Andrew Poulton, Andrew Ryan, Ankit Ramchandani, Annie Franco, Aparajita Saraf, Arkabandhu Chowdhury, Ashley Gabriel, Ashwin Bharambe, Assaf Eisenman, Azadeh Yazdan, Beau James, Ben Maurer, Benjamin Leonhardi, Bernie Huang, Beth Loyd, Beto De~Paola, Bhargavi Paranjape, Bing Liu, Bo~Wu, Boyu Ni, Braden Hancock, Bram Wasti, Brandon Spence, Brani Stojkovic, Brian Gamido, Britt
  Montalvo, Carl Parker, Carly Burton, Catalina Mejia, Changhan Wang, Changkyu Kim, Chao Zhou, Chester Hu, Ching-Hsiang Chu, Chris Cai, Chris Tindal, Christoph Feichtenhofer, Damon Civin, Dana Beaty, Daniel Kreymer, Daniel Li, Danny Wyatt, David Adkins, David Xu, Davide Testuggine, Delia David, Devi Parikh, Diana Liskovich, Didem Foss, Dingkang Wang, Duc Le, Dustin Holland, Edward Dowling, Eissa Jamil, Elaine Montgomery, Eleonora Presani, Emily Hahn, Emily Wood, Erik Brinkman, Esteban Arcaute, Evan Dunbar, Evan Smothers, Fei Sun, Felix Kreuk, Feng Tian, Firat Ozgenel, Francesco Caggioni, Francisco Guzm{\'a}n, Frank Kanayet, Frank Seide, Gabriela~Medina Florez, Gabriella Schwarz, Gada Badeer, Georgia Swee, Gil Halpern, Govind Thattai, Grant Herman, Grigory Sizov, Guangyi, Zhang, Guna Lakshminarayanan, Hamid Shojanazeri, Han Zou, Hannah Wang, Hanwen Zha, Haroun Habeeb, Harrison Rudolph, Helen Suk, Henry Aspegren, Hunter Goldman, Ibrahim Damlaj, Igor Molybog, Igor Tufanov, Irina-Elena Veliche, Itai Gat, Jake
  Weissman, James Geboski, James Kohli, Japhet Asher, Jean-Baptiste Gaya, Jeff Marcus, Jeff Tang, Jennifer Chan, Jenny Zhen, Jeremy Reizenstein, Jeremy Teboul, Jessica Zhong, Jian Jin, Jingyi Yang, Joe Cummings, Jon Carvill, Jon Shepard, Jonathan McPhie, Jonathan Torres, Josh Ginsburg, Junjie Wang, Kai Wu, Kam~Hou U, Karan Saxena, Karthik Prasad, Kartikay Khandelwal, Katayoun Zand, Kathy Matosich, Kaushik Veeraraghavan, Kelly Michelena, Keqian Li, Kun Huang, Kunal Chawla, Kushal Lakhotia, Kyle Huang, Lailin Chen, Lakshya Garg, Lavender A, Leandro Silva, Lee Bell, Lei Zhang, Liangpeng Guo, Licheng Yu, Liron Moshkovich, Luca Wehrstedt, Madian Khabsa, Manav Avalani, Manish Bhatt, Maria Tsimpoukelli, Martynas Mankus, Matan Hasson, Matthew Lennie, Matthias Reso, Maxim Groshev, Maxim Naumov, Maya Lathi, Meghan Keneally, Michael~L. Seltzer, Michal Valko, Michelle Restrepo, Mihir Patel, Mik Vyatskov, Mikayel Samvelyan, Mike Clark, Mike Macey, Mike Wang, Miquel~Jubert Hermoso, Mo~Metanat, Mohammad Rastegari, Munish
  Bansal, Nandhini Santhanam, Natascha Parks, Natasha White, Navyata Bawa, Nayan Singhal, Nick Egebo, Nicolas Usunier, Nikolay~Pavlovich Laptev, Ning Dong, Ning Zhang, Norman Cheng, Oleg Chernoguz, Olivia Hart, Omkar Salpekar, Ozlem Kalinli, Parkin Kent, Parth Parekh, Paul Saab, Pavan Balaji, Pedro Rittner, Philip Bontrager, Pierre Roux, Piotr Dollar, Polina Zvyagina, Prashant Ratanchandani, Pritish Yuvraj, Qian Liang, Rachad Alao, Rachel Rodriguez, Rafi Ayub, Raghotham Murthy, Raghu Nayani, Rahul Mitra, Raymond Li, Rebekkah Hogan, Robin Battey, Rocky Wang, Rohan Maheswari, Russ Howes, Ruty Rinott, Sai~Jayesh Bondu, Samyak Datta, Sara Chugh, Sara Hunt, Sargun Dhillon, Sasha Sidorov, Satadru Pan, Saurabh Verma, Seiji Yamamoto, Sharadh Ramaswamy, Shaun Lindsay, Shaun Lindsay, Sheng Feng, Shenghao Lin, Shengxin~Cindy Zha, Shiva Shankar, Shuqiang Zhang, Shuqiang Zhang, Sinong Wang, Sneha Agarwal, Soji Sajuyigbe, Soumith Chintala, Stephanie Max, Stephen Chen, Steve Kehoe, Steve Satterfield, Sudarshan
  Govindaprasad, Sumit Gupta, Sungmin Cho, Sunny Virk, Suraj Subramanian, Sy~Choudhury, Sydney Goldman, Tal Remez, Tamar Glaser, Tamara Best, Thilo Kohler, Thomas Robinson, Tianhe Li, Tianjun Zhang, Tim Matthews, Timothy Chou, Tzook Shaked, Varun Vontimitta, Victoria Ajayi, Victoria Montanez, Vijai Mohan, Vinay~Satish Kumar, Vishal Mangla, V{\'i}tor Albiero, Vlad Ionescu, Vlad Poenaru, Vlad~Tiberiu Mihailescu, Vladimir Ivanov, Wei Li, Wenchen Wang, Wenwen Jiang, Wes Bouaziz, Will Constable, Xiaocheng Tang, Xiaofang Wang, Xiaojian Wu, Xiaolan Wang, Xide Xia, Xilun Wu, Xinbo Gao, Yanjun Chen, Ye~Hu, Ye~Jia, Ye~Qi, Yenda Li, Yilin Zhang, Ying Zhang, Yossi Adi, Youngjin Nam, Yu, Wang, Yuchen Hao, Yundi Qian, Yuzi He, Zach Rait, Zachary DeVito, Zef Rosnbrick, Zhaoduo Wen, Zhenyu Yang, and Zhiwei Zhao.
\newblock The {{Llama}} 3 {{Herd}} of {{Models}}, August 2024.

\bibitem{gpt4o}
OpenAI Team.
\newblock Hello {{GPT-4o}} {\textbar} {{OpenAI}}.
\newblock https://openai.com/index/hello-gpt-4o/.

\bibitem{IntroducingOpenAIo1_2024sep}
Introducing {{OpenAI}} o1.
\newblock https://openai.com/o1/, September 2024.

\bibitem{bonnefon_2018mar}
Jean-Fran{\c c}ois Bonnefon.
\newblock The {{Pros}} and {{Cons}} of {{Identifying Critical Thinking}} with {{System}} 2 {{Processing}}.
\newblock {\em Topoi}, 37(1):113--119, March 2018.

\bibitem{bostrom_2018}
Nick Bostrom, Allan Dafoe, and Carrick Flynn.
\newblock Public {{Policy}} and {{Superintelligent AI}}: {{A Vector Field Approach}}.
\newblock {\em Ethics of Artificial Intelligence}, 2018.

\bibitem{rae_2022jan}
Jack~W. Rae, Sebastian Borgeaud, Trevor Cai, Katie Millican, Jordan Hoffmann, Francis Song, John Aslanides, Sarah Henderson, Roman Ring, Susannah Young, Eliza Rutherford, Tom Hennigan, Jacob Menick, Albin Cassirer, Richard Powell, George van~den Driessche, Lisa~Anne Hendricks, Maribeth Rauh, Po-Sen Huang, Amelia Glaese, Johannes Welbl, Sumanth Dathathri, Saffron Huang, Jonathan Uesato, John Mellor, Irina Higgins, Antonia Creswell, Nat McAleese, Amy Wu, Erich Elsen, Siddhant Jayakumar, Elena Buchatskaya, David Budden, Esme Sutherland, Karen Simonyan, Michela Paganini, Laurent Sifre, Lena Martens, Xiang~Lorraine Li, Adhiguna Kuncoro, Aida Nematzadeh, Elena Gribovskaya, Domenic Donato, Angeliki Lazaridou, Arthur Mensch, Jean-Baptiste Lespiau, Maria Tsimpoukelli, Nikolai Grigorev, Doug Fritz, Thibault Sottiaux, Mantas Pajarskas, Toby Pohlen, Zhitao Gong, Daniel Toyama, Cyprien de~Masson {d'Autume}, Yujia Li, Tayfun Terzi, Vladimir Mikulik, Igor Babuschkin, Aidan Clark, Diego de~Las Casas, Aurelia Guy, Chris
  Jones, James Bradbury, Matthew Johnson, Blake Hechtman, Laura Weidinger, Iason Gabriel, William Isaac, Ed~Lockhart, Simon Osindero, Laura Rimell, Chris Dyer, Oriol Vinyals, Kareem Ayoub, Jeff Stanway, Lorrayne Bennett, Demis Hassabis, Koray Kavukcuoglu, and Geoffrey Irving.
\newblock Scaling {{Language Models}}: {{Methods}}, {{Analysis}} \& {{Insights}} from {{Training Gopher}}, January 2022.

\bibitem{wei_2023jan}
Jason Wei, Xuezhi Wang, Dale Schuurmans, Maarten Bosma, Brian Ichter, Fei Xia, Ed~Chi, Quoc Le, and Denny Zhou.
\newblock Chain-of-{{Thought Prompting Elicits Reasoning}} in {{Large Language Models}}, January 2023.

\bibitem{kojima_2023jan}
Takeshi Kojima, Shixiang~Shane Gu, Machel Reid, Yutaka Matsuo, and Yusuke Iwasawa.
\newblock Large {{Language Models}} are {{Zero-Shot Reasoners}}, January 2023.

\bibitem{yao_2023may}
Shunyu Yao, Dian Yu, Jeffrey Zhao, Izhak Shafran, Thomas~L. Griffiths, Yuan Cao, and Karthik Narasimhan.
\newblock Tree of {{Thoughts}}: {{Deliberate Problem Solving}} with {{Large Language Models}}, May 2023.

\bibitem{besta_2023nov}
Maciej Besta, Nils Blach, Ales Kubicek, Robert Gerstenberger, Lukas Gianinazzi, Joanna Gajda, Tomasz Lehmann, Michal Podstawski, Hubert Niewiadomski, Piotr Nyczyk, and Torsten Hoefler.
\newblock Graph of {{Thoughts}}: {{Solving Elaborate Problems}} with {{Large Language Models}}, November 2023.

\bibitem{lee_2023jun}
Soochan Lee and Gunhee Kim.
\newblock Recursion of {{Thought}}: {{A Divide-and-Conquer Approach}} to {{Multi-Context Reasoning}} with {{Language Models}}, June 2023.

\bibitem{ning_2023oct}
Xuefei Ning, Zinan Lin, Zixuan Zhou, Zifu Wang, Huazhong Yang, and Yu~Wang.
\newblock Skeleton-of-{{Thought}}: {{Large Language Models Can Do Parallel Decoding}}, October 2023.

\bibitem{chen_2023octb}
Wenhu Chen, Xueguang Ma, Xinyi Wang, and William~W. Cohen.
\newblock Program of {{Thoughts Prompting}}: {{Disentangling Computation}} from {{Reasoning}} for {{Numerical Reasoning Tasks}}, October 2023.

\bibitem{wang_2023mar}
Xuezhi Wang, Jason Wei, Dale Schuurmans, Quoc Le, Ed~Chi, Sharan Narang, Aakanksha Chowdhery, and Denny Zhou.
\newblock Self-{{Consistency Improves Chain}} of {{Thought Reasoning}} in {{Language Models}}, March 2023.

\bibitem{chia_2023nov}
Yew~Ken Chia, Guizhen Chen, Luu~Anh Tuan, Soujanya Poria, and Lidong Bing.
\newblock Contrastive {{Chain-of-Thought Prompting}}, November 2023.

\bibitem{diao_2023may}
Shizhe Diao, Pengcheng Wang, Yong Lin, and Tong Zhang.
\newblock Active {{Prompting}} with {{Chain-of-Thought}} for {{Large Language Models}}, May 2023.

\bibitem{wang_2022jul}
Xuezhi Wang, Jason Wei, Dale Schuurmans, Quoc Le, Ed~Chi, and Denny Zhou.
\newblock Rationale-{{Augmented Ensembles}} in {{Language Models}}, July 2022.

\bibitem{zhang_2022oct}
Zhuosheng Zhang, Aston Zhang, Mu~Li, and Alex Smola.
\newblock Automatic {{Chain}} of {{Thought Prompting}} in {{Large Language Models}}, October 2022.

\bibitem{yang_2018sep}
Zhilin Yang, Peng Qi, Saizheng Zhang, Yoshua Bengio, William~W. Cohen, Ruslan Salakhutdinov, and Christopher~D. Manning.
\newblock {{HotpotQA}}: {{A Dataset}} for {{Diverse}}, {{Explainable Multi-hop Question Answering}}, September 2018.

\bibitem{geva_2021}
Mor Geva, Daniel Khashabi, Elad Segal, Tushar Khot, Dan Roth, and Jonathan Berant.
\newblock Did {{Aristotle Use}} a {{Laptop}}? {{A Question Answering Benchmark}} with {{Implicit Reasoning Strategies}}.
\newblock {\em Transactions of the Association for Computational Linguistics}, 9:346--361, 2021.

\bibitem{chen_2023octd}
Lingjiao Chen, Matei Zaharia, and James Zou.
\newblock How is {{ChatGPT}}'s behavior changing over time?, October 2023.

\end{thebibliography}

\pagebreak

\appendix
\section{Appendix}

\subsection{AutoReason Prompt Template}

\begin{lstlisting}
export const autoReasonPrompt = ({ question }: { question: string }) => {
  return `You will formulate Chain of Thought (CoT) reasoning traces.
  CoT is a prompting technique that helps you to think about a problem in a structured way.
  It breaks down a problem into a series of logical reasoning traces.
  
  You will be given a question and using this question you will decompose the question into a series of logical reasoning traces.
  Only write the reasoning traces and do not answer the question yourself.
  
  Here are some examples of CoT reasoning traces:
  
  Question: Did Brazilian jiu-jitsu Gracie founders have at least a baker's dozen of kids between them?
  
  Reasoning traces:
  - Who were the founders of Brazilian jiu-jitsu?
  - What is the number represented by the baker's dozen?
  - How many children do Gracie founders have altogether
  - Is this number bigger than baker's dozen?
  
  Question: Is cow methane safer for environment than cars
  
  Reasoning traces:
  - How much methane is produced by cars annually?
  - How much methane is produced by cows annually?
  - Is methane produced by cows less than methane produced by cars?
  
  Question: ${question}
  
  Reasoning traces:
  `;
};
\end{lstlisting}

\subsection{HotpotQA Base Prompt}

\begin{lstlisting}
export const baseHotpotqaPrompt = `You're an agent. Your job is to answer some questions. Here are the rules:
1. You will be given a question
2. You will answer the question with a short answer, it might yes/no or a short phrase
3. When you know the answer, write it in this format only: "<answer>"`;
\end{lstlisting}

\pagebreak

\subsection{HotpotQA CoT Prompt}

\begin{lstlisting}
export const cotHotpotQaPrompt = ({ question }: { question: string }) => {
  return `Your job is to answer some questions. Here are some examples of how you should answer:
  
  Q: Do hamsters provide food for any animals?
  Hamsters are prey animals. Prey are food for predators. Thus, hamsters provide food for some animals.
  Answer: yes
  
  Q: Could Brooke Shields succeed at University of Pennsylvania?
  Brooke Shields went to Princeton University. Princeton University is about as academically rigorous as the University of Pennsylvania. Thus, Brooke Shields could also succeed at the University of Pennsylvania.
  Answer: yes
  
  Q: Yes or no: Hydrogen's atomic number squared exceeds number of Spice Girls?
  "Hydrogen has an atomic number of 1. 1 squared is 1. There are 5 Spice Girls. Thus, Hydrogen's atomic number squared is less than 5.
  Answer: no
  
  Q: ${question}
  `;
};

\end{lstlisting}

\subsection{StrategyQA Base Prompt}

\begin{lstlisting}
export const baseStrategyQaPrompt = `You're an agent. Your job is to answer some questions. Here are the rules:
1. You will be given a question
2. You will answer the question with true or false
3. When you know the answer, write it in this format only: "answer"`;
\end{lstlisting}

\subsection{StrategyQA CoT Prompt}

\begin{lstlisting}
export const cotStrategyQaPrompt = ({ question }: { question: string }) => {
  return `Your job is to answer some questions. Here are some examples of how you should answer:
  
  Q: Do hamsters provide food for any animals?
  Hamsters are prey animals. Prey are food for predators. Thus, hamsters provide food for some animals.
  Answer: yes
  
  Q: Could Brooke Shields succeed at University of Pennsylvania?
  Brooke Shields went to Princeton University. Princeton University is about as academically rigorous as the University of Pennsylvania. Thus, Brooke Shields could also succeed at the University of Pennsylvania.
  Answer: yes
  
  Q: Yes or no: Hydrogen's atomic number squared exceeds number of Spice Girls?
  "Hydrogen has an atomic number of 1. 1 squared is 1. There are 5 Spice Girls. Thus, Hydrogen's atomic number squared is less than 5.
  Answer: no
  
  Q: ${question}
  `;
};

\end{lstlisting}

\pagebreak

\subsection{Scorer Prompt}

\begin{lstlisting}
export const scorePrompt = ({
  question,
  answer,
  correctAnswer,
}: ScorePromptParameters) => {
  return `Your job is to score an answer's correctness from 0 to 10. You will be given the question, the correct answer, and the answer you need to score.
  0 means the answer is completely wrong, 10 means the answer is completely correct. Explain your reasoning first shortly, and then write the score as a literal number (0 to 10).

  Question: ${question}
  Answer: ${answer}
  Correct Answer: ${correctAnswer}
  Score: `;
};

\end{lstlisting}

\end{document}